\documentclass[11pt,a4paper]{article}
\usepackage[hyperref]{emnlp2018}
\usepackage{times}
\usepackage{url}
\usepackage{latexsym}
\usepackage{lipsum}
\usepackage{mathtools}
\usepackage{amssymb}
\usepackage{hyperref}
\usepackage{array}
\usepackage{float}

\usepackage{url}

\aclfinalcopy % Uncomment this line for the final submission
\title{Co-Stack Residual Affinity Networks with Multi-level Attention Refinement for Matching Text Sequences}

\author{Yi Tay$^\dagger$\thanks{\:\:Denotes equal contribution.}\:, Luu Anh Tuan$^\psi$\footnotemark[1]\:, Siu Cheung Hui$^\phi$ \\
  $^\dagger$$^\phi$Nanyang Technological University, Singapore \\
  $^\psi$Institute for Infocomm Research, A*Star Singapore \\
  {\tt ytay017@e.ntu.edu.sg$^\dagger$, at.luu@i2r.a-star.edu.sg$^\psi$} \\
  {\tt asschui@ntu.edu.sg$^\phi$} \\
  }

\date{}

\begin{document}
\maketitle
\begin{abstract}
Learning a matching function between two text sequences is a long standing problem in NLP research. This task enables many potential applications such as question answering and paraphrase identification. This paper proposes Co-Stack Residual Affinity Networks (CSRAN), a new and universal neural architecture for this problem. CSRAN is a deep architecture, involving stacked (multi-layered) recurrent encoders. Stacked/Deep architectures are traditionally difficult to train, due to the inherent weaknesses such as difficulty with feature propagation and vanishing gradients. CSRAN incorporates two novel components to take advantage of the stacked architecture. Firstly, it introduces a new bidirectional alignment mechanism that learns affinity weights by fusing sequence pairs across stacked hierarchies. Secondly, it leverages a multi-level attention refinement component between stacked recurrent layers. The key intuition is that, by leveraging information across all network hierarchies, we
can not only improve gradient flow but also improve overall performance. We conduct extensive experiments on six well-studied text sequence matching datasets, achieving state-of-the-art performance on all.
\end{abstract}

\section{Introduction}
Determining the semantic affinity between two text sequences is a long standing research problem in natural language processing research. This is understandable, given that technical innovations in this domain would naturally bring benefits to a diverse plethora of applications ranging from paraphrase detection to standard document retrieval. This work focuses on short textual sequences, focusing on a myriad of applications such as natural language inference, question answering, reply prediction and paraphrase detection. This paper presents a new deep matching model for universal text matching.

Neural networks are dominant state-of-the-art approaches for many of these matching problems \cite{DBLP:journals/corr/abs-1709-04348,shen2017inter,1702.03814,DBLP:conf/acl/ChenZLWJI17}. Fundamentally, neural networks operate via a concept of feature hierarchy, in which hierarchical representations are constructed as sequences propagate across the network. In the context of matching, representations are often (1) encoded, (2) matched, and then (3) aggregated for prediction. Each key step often comprises several layers, which consequently adds to the overall depth of the network.

Unfortunately, it is a well established fact that deep networks are difficult to train. This is attributed to not only vanishing/exploding gradients but also
an instrinsic difficulty pertaining to feature propagation. To this end,
commonly adopted solutions include Residual connections \cite{he2016deep} and/or Highway layers \cite{DBLP:journals/corr/SrivastavaGS15}. The key idea in these approaches is to introduce additional (skip/residual) connections, propagating shallower layers to deeper layers via shortcuts. To the best of our knowledge, these techniques are generally applied to single sequences and therefore the notion of pairwise residual connections have not been explored.

This paper presents Co-Stack Residual Affinity Networks (CSRAN), a stacked multi-layered recurrent architecture for general purpose text matching.  Our model proposes a new co-stacking mechanism that computes bidirectional affinity scores by leveraging all feature hierarchies between text sequence pairs. More concretely, word-by-word affinity scores are not computed just from the final encoded representations but across all the entire feature hierarchy.

There are several benefits to our co-stacking mechanism. Firstly, co-stacking acts as a form of residual connector, alleviating the instrinsic issues with network depth. Secondly, there are more extensive matching interfaces between text sequences as the affinity matrix is not computed by just one representation but multiple representations instead. Naturally, increasing the opportunities for interactions between sequences is an intuitive method for improving performance.

 Additionally, our model incorporates a Multi-level Attention Refinement (MAR) architecture in order to fully leverage the stacked recurrent architecture. The MAR architecture is a multi-layered adaptation and extension of the CAFE model \cite{tay2017compare}, in which attention is computed, compressed and then re-fed into the input sequence. In our approach, we use CAFE blocks to repeatedly \textit{refine} representations at each level of the stacked recurrent encoder.

 The overall outcome of the above-mentioned architectural synergies is a highly competitive model that establishes state-of-the-art performance on six well-known text matching datasets such as SNLI and TrecQA. The overall contributions of this work are summarized as follows:
\begin{itemize}
\item We propose a new deep stacked recurrent architecture for matching text sequences. Our model is based on a new co-stacking mechanism which learns to align by exploiting matching across feature hierarchies. This can be interpreted as a new way to incorporate shortcut connections within neural models for sequence matching. Additionally, we also propose a multi-level attention refinement scheme to leverage our stacked recurrent model.
\item While stacked architectures can potentially lead to considerable improvements in performance, our experiments show that in the absence of our proposed CSRA (Co-stack Residual Affinity) mechanism, stacking may conversely lead to performance degradation. As such, this demonstrates that our proposed techniques are essential for harnessing the potential of deep architectures.
\item We conduct extensive experiments on four text matching tasks across six well-studied datasets, i.e., \textit{Natural Language Inference} (SNLI \cite{DBLP:conf/emnlp/BowmanAPM15}, SciTail \cite{scitail}), \textit{Paraphrase Identification} (Quora, TwitterURL \cite{lan2017continuously}), \textit{Answer Sentence Selection} \cite{DBLP:conf/emnlp/WangSM07} and \textit{Utterance-Response Matching} (Ubuntu \cite{lowe2015ubuntu}).  Our model achieves state-of-the-art performance
on all datasets.
\end{itemize}

% \subsection{Our Contributions}

% In many of these models, representations are typically encoded, matched and then aggregated. Each functional segment of the neural model easily comprises several layers. For example, it is common to utilize stacked recurrent encoders \cite{DBLP:conf/repeval/NieB17} or multi-layered neural networks (or highway networks) \cite{tay2017compare} for encoding representations. While there are inherent benefits to going deep, this is often prohibitive due to its instrinsic weaknessess.

\section{Co-Stack Residual Affinity Networks}

\begin{figure*}
  \centering
    \includegraphics[width=0.85\linewidth]{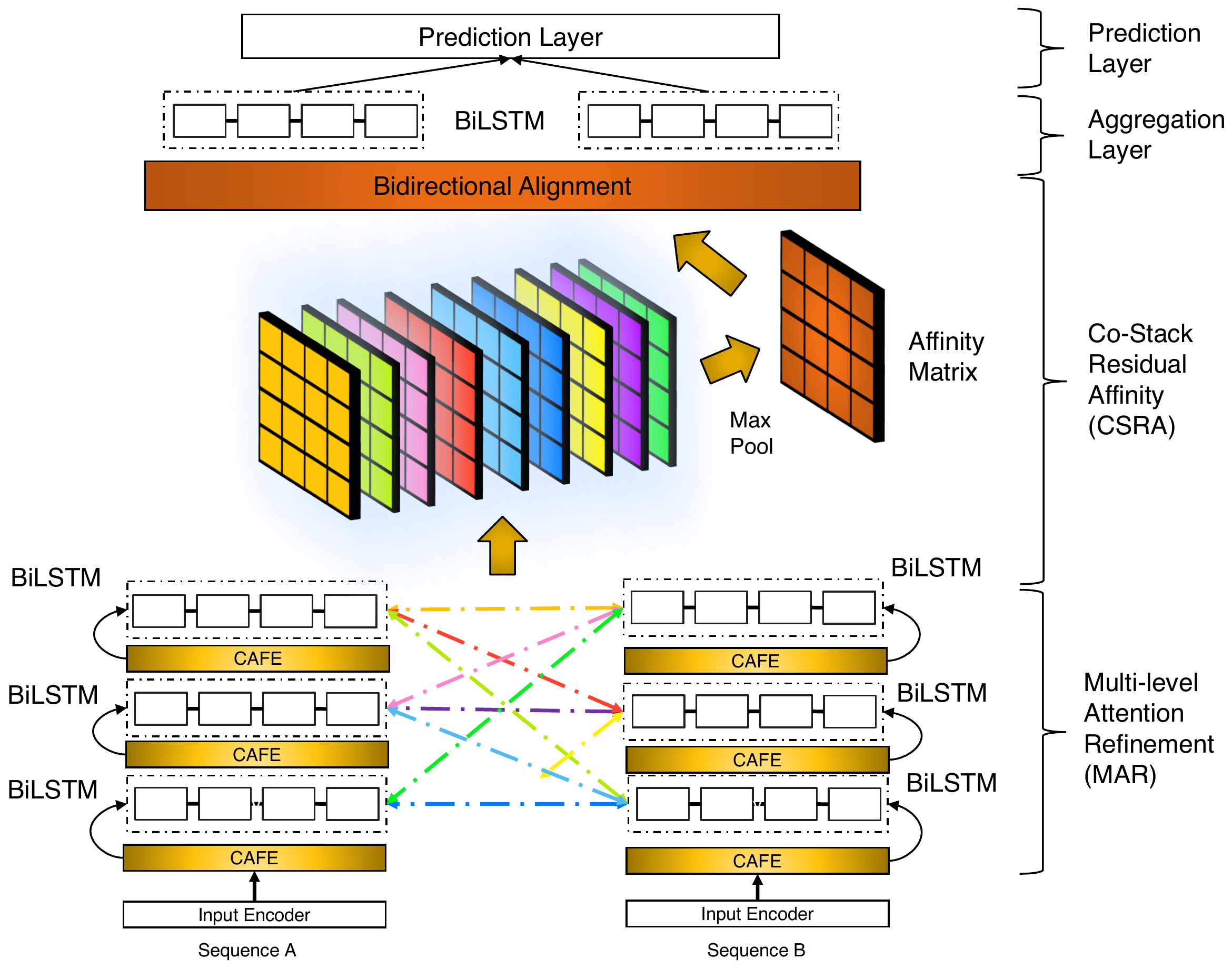}
    \caption{Illustration of the proposed Co-Stack Residual Affinity Network (CSRAN) architecture. Each color coded matrix represents the interactions between two layers of sequence A and sequence B. (Best viewed in color)}
    \label{fig:csra}
\end{figure*}

In this section, we introduce our proposed model architecture for general/universal text matching. The key idea of this architecture is to
leverage deep stacked layers, while mitigating the inherent weaknesses of going deep. As such, our network is in similar spirit to highway networks, residual networks and DenseNets, albeit tailored specifically for pairwise architectures. Figure \ref{fig:csra} illustrates a high-level overview of our proposed model architecture.

\subsection{Input Encoder}
The inputs to our model are standard sequences of words $A$ and $B$ which
represent sequence $a$ and sequence $b$ respectively. In the context of different applications, $a$ and $b$ take different roles such as premise/hypothesis
or question/answer. Both sequences are converted into word representations (via pretrained word embeddings) and
character-based representations. Character embeddings are trainable parameters and a final character-based word representation of $d$ dimensions is learned by
passing all characters into a Bidirectional LSTM encoder. This is standard, following many works such as \cite{1702.03814}. Word embeddings and character-based word representations are then concatenated to form the final
word representation.  Next, the word representation is passed through a (optional and tuned as a hyperparameter) 2-layered highway network of $d$ dimensions.
\subsection{Stacked Recurrent Encoders}
Next, word representations are passed into a stacked recurrent encoder layer. Specifically, we use Bidirectional LSTM encoders at this layer. Let $k$ be the number of
layers of the stacked recurrent encoder layer.
\begin{align}
h^{i}_{t} = \text{BiLSTM}_{i}(h_{t-1}) \:\: \forall t \in [1,2 \cdots \ell]
\end{align}
where BiLSTM$_i$ represents the i-th BiLSTM layer and $h^{i}_{t}$ represents
the t-th hidden state of the i-th BiLSTM layer. $\ell$ is the sequence length. Note that the parameters are shared for both $a$ and $b$.

% The final output of $h_{t}$ is the concatenation of $[h^{1}_{t};h^{2}_{t}\cdots h^{k}_{t}]$. This exposes the inner atte

\subsection{Multi-level Attention Refinement (MAR)}

Inspired by CAFE \cite{tay2017compare} (Compare-Align-Factorized Encoders), a top performing model on the SNLI benchmark, we utilize CAFE blocks between the BiLSTM layers. Each CAFE block returns \textit{six} features, which are generated by a factorization operation using factorization machines (FM). While the authors in \cite{tay2017compare} simply use this operation in a single layer, we utilize this in a multi-layered fashion which we found to have worked well. This constitutes our multi-level attention refinement mechanism.
More concretely, we apply the CAFE operation to the outputs of each BiLSTM layer, allowing the next BiLSTM layer to process the `augmented' representations. The next layer retains its dimensionality by projecting the augmented representation back to its original size using the BiLSTM encoder. This can be interpreted as repeatedly refining representations via attention. As such, adding CAFE blocks is a very natural fit to the stacked recurrent architecture.

\paragraph{CAFE Blocks}
This section describes the operation of each CAFE block. The key idea behind CAFE blocks is to align $a$ and $b$, and compress alignment vectors such as $b'-a$ (subtraction), $b' \odot a$ (element-wise multiplication) and $[b';a]$ (concatenation) into scalar features. These scalar features are concatenated to the original input embedding, which can be pased into another BiLSTM layer for refining representations. Firstly, $a, b$ are modeled aligned via $E_{ij} = F(a)^{\top}F(b)$ and then aligned via:
\begin{align}
A' = E^{\top}B \:\: \text{and} \:\: B' = AE^{\top}
\end{align}
Given aligned pairs $(A',B)$ and $(B',A)$, we generate \textit{three} matching vectors for the concatenation ($[a'_i; b_i]$), element-wise multiplication ($a'_i \odot b_i$) and subtraction vectors ($a'_i - b_i$) of each pair. After which, we apply a factorization machine \cite{rendle2010factorization}  $M(x)$ on each matching vector.
\begin{align}
M(x) &= w_{0} + \sum^{n}_{i=1} w_i \: x_i  + \sum^{n}_{i=1} \sum^{n}_{j=i+1} \langle v_i, v_j \rangle \: x_i \:x_j
\end{align}
where $v \in \mathbb{R}^{d \times k}$, $w_{0} \in \mathbb{R}$ and $w_{i} \in \mathbb{R}^{d}$. The output $M(x)$ is a scalar. Intuitively, this layer tries to learn pairwise interactions between every $x_i$ and $x_j$ using factorized (vector) parameters $v$.

Factorization machines model low-rank structure within the matching vector, producing a scalar feature. This enables efficient propagation of these matching features to the next layer. The output of each CAFE block is the original input to the CAFE module, augmented with the output of the factorization machines. As such, if the input sequence is of $d$ dimensions, then the output is $d+3$ dimensions. Additionally, intra-attention is applied in similar fashion as above to generate \textit{three} more features for each sequence. As a result, the output dimensions for each word becomes $d+6$.

\subsection{Co-Stack Residual Affinity (CSRA)}

This layer is the cornerstone of our proposed approach and is represented as the middle segment of Figure \ref{fig:csra} (the colorful matrices).

\paragraph{Co-Stacking} Co-stacking refers to the fusion of $a$ and $b$ across multiple hierarchies. Recall that the affinity score between two words is typically computed by $s_{ij}=a^{\top}b$.
We extend this to a residual formulation. More concretely, the affinity score
between both words is now computed as the \textit{maximum} influence it has over
all layers.
\begin{align}
s_{ij} = \max \sum_{p} \sum_{q} a_{pi}^{\top}\:b_{qj}
\end{align}
where $a_{pi}$ is the i-th word for the p-th stacked layer for $a$ and $b_{qj}$ is the
j-th word for the q-th stacked layer for $b$. The choice of the maximum operator is
intuitive and is strongly motivated by the fact that we would like to give a high affinity for each word pair that shows a strong match at any of different
hierarchical stages of learning representations. Note that this layer can be
interpreted as constructing a matching tensor based on multi-hierarchical information and selecting the most informative match across all representation hierarchies.

\paragraph{Bidirectional Alignment}
In order to learn (bidirectionally) attentive representations, we first concatenate all stacked outputs to form a $\ell \times kd$ vector. Next, we apply
the following operations to $A \in \mathbb{R}^{\ell_{a} \times kd}$ and $B \in \mathbb{R}^{\ell_{b} \times kd}$.
\begin{align}
\bar{A} = S^{\top}B \:\: \text{and} \:\: \bar{B} = AS^{\top}
\end{align}
where $\bar{A} \in \mathbb{R}^{\ell_{b} \times kd}, \bar{B} \in \mathbb{R}^{\ell_{a} \times kd}$ are the attentive (aligned) representations.
\subsubsection{Matching and Aggregation Layer}
Next, we match the attentive (aligned) representations using the subtraction,
element-wise multiplication and concatenation of each aligned word. Subsequently, we pass this matching vector into a $k$ layered BiLSTM layer.
\begin{align}
a'_{i} = \text{BiLSTM}_k([\bar{b}_{i}-a_{i}, \bar{b}_{i} \odot a_{i}, \bar{b}_{i}, a_{i}]) \\
b'_{i} = \text{BiLSTM}_k([\bar{a}_{i}-b_{i}, \bar{a}_{i} \odot b_{i}, \bar{a}_{i}, b_{i}])
\end{align}
The final feature representation is learned via the summation across the temporal dimension as follows:
\begin{align}
z = [\sum_{i=1}^{\ell_{a}} a'_{i} \:\:;\:\: \sum_{i=1}^{\ell_{b}}  b'_{i}]
\end{align}
where $[.;.]$ is the concatenation operator.
\subsection{Output and Prediction Layer}
Our model predicts using the feature vector $z$ for every given sequence pair.
At this layer, we utilize standard fully connected layers. The number of  output layers is typically 2-3 and is a tuned hyperparameter. Softmax is applied onto the final layer. The final layer is application specific, e.g., $k$ classes for classification tasks and a two-class softmax for pointwise ranking. For all datasets, we optimize the cross entropy loss.

\section{Experimental Evaluation}
In this section, we introduce our experimental setup, baselines and results.
\subsection{Datasets and Competitor Baselines}
We use six public benchmark datasets for evaluating our proposed approach. This section briefly introduces each dataset, along with several state-of-the-art approaches that we compare against. Table \ref{tab:dataset} provides a summary of the datasets used in our experiments.

% Table generated by Excel2LaTeX from sheet 'Sheet4'
\begin{table}[htbp]
  \centering

    \begin{tabular}{llcc}
      \hline
    Dataset & Task  &  $|C|$     & Pairs \\
    \hline
    SNLI  & Premise-Hypothesis & 3 & 570K \\
    Scitail & Premise-Hypothesis & 2 & 27K \\
    Quora & Question-Question & 2 & 400K \\
    Twitter & Tweet-Tweet & 2 & 51K \\
    TrecQA & Question-Answer & R  & 56K \\
    Ubuntu & Utterance-Response & R  & 1M \\
    \hline
    \end{tabular}%
      \caption{Statistics of datasets used in our experiment. $|C|$ denotes the number of classes and R denotes a ranking formulation. Twitter stands for the TwitterURL dataset.}
  \label{tab:dataset}%
\end{table}%

\paragraph{Stanford Natural Language Inference (SNLI)} \cite{DBLP:conf/emnlp/BowmanAPM15} is a well-known dataset for
\textit{entailment classification} (or natural language inference). The task is to determine if two sequences entail/contradict or are neutral to each other. This task is a three-way classification problem. On this dataset, we compare with several state-of-the-art models such as BiMPM \cite{1702.03814}, ESIM \cite{DBLP:conf/acl/ChenZLWJI17}, DIIN \cite{DBLP:journals/corr/abs-1709-04348}, DR-BiLSTM \cite{ghaeini2018dr} and CAFE \cite{tay2017compare}.
\paragraph{Science Entailment (SciTail)} \cite{scitail} is a new entailment classification dataset that was constructed from science questions and answers. This dataset involves two-way classification (entail or non-entail). We compare with DecompAtt \cite{DBLP:conf/emnlp/ParikhT0U16}, ESIM, DGEM \cite{scitail} and CAFE.
\paragraph{Quora Duplicate Detection} is a well-studied paraphrase identification dataset\footnote{\url{https://data.quora.com/First-Quora-Dataset-Release-Question-Pairs}}. We use the splits provided by \cite{1702.03814}. The task is to determine if two questions are paraphrases of each other. This task is formulated as a binary classication problem. We compare with L.D.C \cite{wang2016sentence}, BiMPM, the DecompAtt implementation by \cite{tomar2017neural} (word and char level) and DIIN.
\paragraph{TwitterURL} \cite{lan2017continuously} is another dataset for paraphrase identification. It was constructed using Tweets referring to news articles. This task is also a binary classification problem. We compare with (1) MultiP \cite{xu2014extracting}, a strong baseline, (2) the implementation of \cite{DBLP:conf/naacl/HeL16} by \cite{lan2017continuously} and (3) the Subword + LM model from \cite{lan2018subword}.
\paragraph{TrecQA} \cite{DBLP:conf/emnlp/WangSM07} is a well-studied dataset for \textit{answer sentence selection task} (or question-answer matching). The goal is to rank answers given a question. This task is formulated as a pointwise learning-to-rank problem. Baselines include HyperQA \cite{DBLP:journals/corr/TayLH17a}, Ranking-based Multi-Perspective CNN \cite{DBLP:conf/emnlp/HeGL15} implementation by \cite{DBLP:conf/cikm/RaoHL16}, BiMPM, the compare-aggregate \cite{DBLP:journals/corr/WangJ16b} model extension by \cite{Bian:2017:CMD:3132847.3133089} (we denote this model as CA), IWAN \cite{shen2017inter} and the recent MCAN model, i.e., Multi-Cast Attention Networks \cite{Tay:2018:MAN:3219819.3220048}. A leaderboard is maintained at \url{https://aclweb.org/aclwiki/Question_Answering_(State_of_the_art)}.
\paragraph{Ubuntu} \cite{lowe2015ubuntu} is a dataset for \textit{Utterance-Response Matching} and comprises 1-million utterance-response pairs. This dataset is based on the Ubuntu dialogue corpus. The goal is to predict the response to a message. We use the same setup as \cite{wu2016knowledge}. Baselines include CNTN \cite{DBLP:conf/ijcai/QiuH15}, APLSTM \cite{DBLP:journals/corr/SantosTXZ16},
MV-LSTM \cite{DBLP:conf/aaai/WanLGXPC16} and KEHNN \cite{wu2016knowledge}. Results are reported from \cite{wu2016knowledge}.

\paragraph{Metrics} For all datasets, we follow the evaluation procedure from all the original papers. The metric for SNLI, SciTail and Quora is the accuracy metric. The metric for the TwitterURL dataset is the F1 score. The metric for TrecQA is the Mean Average Precision (MAP) and Mean Reciprocal Rank (MRR) metric. The metric for Ubuntu is the Recall@K for $k={1,2,5}$ (given 9 negative samples) and the binary classification accuracy score.
% \subsection{Competitor Methods and Baselines}
% For each dataset, we compare with several state-of-the-art baseline.
% \textbf{SNLI}
\subsection{Experimental Setup}
All baselines are reported from the respective papers. All models are trained with the Adam optimizer \cite{DBLP:journals/corr/KingmaB14} with learning rates tuned amongst $\{0.001, 0.0003,0.0004 \}$. Batch size is tuned amongst $\{32,64,128,256\}$. The dimensions of the BiLSTM encoders are tuned amongst $\{64, 100,200,300\}$ and the number of hidden dimensions of the prediction layers are tuned amongst $\{100,200,300,600\}$. The number of stacked recurrent layers is tuned from $[2,5]$ and the number of aggregation BiLSTM layers is tuned amongst $\{1,2\}$. The number of prediction layers is tuned from [1,3]. Parameters are initialized using glorot uniform \cite{glorot2010understanding}. All unspecified activation functions are ReLU activations. Word embeddings are initialized with GloVe \cite{DBLP:conf/emnlp/PenningtonSM14} and fixed during training. We implement our model in Tensorflow \cite{tensorflow2015-whitepaper} and use the \textsc{cuDNN} implementation for all BiLSTM layers.

\subsection{Experimental Results}
Overall, our proposed CSRAN architecture achieves state-of-the-art performance on all six well-established datasets.

On SNLI (Table \ref{snli}), CSRAN achieves the best\footnote{For fair comparison, we do not compare with (1) models that use external contextualized word embeddings, e.g., CoVe \cite{mccann2017learned} / ELMo \cite{peters2018deep} / generative pretraining \cite{radford2018improving} and (2) ensemble systems. As either (1) and/or (2) would also intuitively boost the performance of the base CSRAN model. \textbf{Update}: After the paper acceptance notification we ran CSRAN + ELMo (embedding layer only) and achieved 89.0 test accuracy.} single model performance to date\footnote{As of EMNLP 2018 Submission} on the well-established dataset. This demonstrates the effectiveness of CSRAN, taking into consideration of the inherent competitiveness of this well-known benchmark. On SciTail (Table \ref{scitail}), CSRAN similarly achieves the best performance to date on this dataset, outperforming the existing CAFE model by $+3.4\%$ absolute accuracy.

On Quora (Table \ref{quora}), CSRAN also achieves the best single model score, outperforming strong baselines such as BiMPM ($+1.1\%$) and DIIN ($+0.2\%$).  Moreover, there is also considerable performance improvement on the TwitterURL dataset (Table \ref{twitterurl}) as CSRAN outperforms the existing state-of-the-art Subword + LM model ($+8\%$) and Deep Pairwise Word ($+9.1\%$).

On TrecQA (Table \ref{trecqa}), CSRAN achieves the best performance on this dataset. CSRAN outperforms the existing state-of-the-art model, IWAN ($+3.2\%/+4.6\%$). CSRAN also outperforms strong competitive baselines such as BiMPM ($+5.2\%/+3.6\%$) and MPCNN ($+5.3\%/+5.8\%$). Finally, on Ubuntu (Table \ref{ubuntu}), CSRAN also outperforms many competitive models such as CNTN, APLSTM and KEHNN. Performance improvement over all metrics are $\approx 9\%-10\%$ compared to the existing state-of-the-art.

Overall, CSRAN achieves state-of-the-art performance on six well-studied datasets. On several datasets, our achieved performance is not only the highest reported score but also outperforms the existing state-of-the-art models by a considerable margin.

\begin{table}[H]
  \centering
\begin{tabular}{lc}
  \hline
Model & Acc \\
\hline
BiMPM \cite{1702.03814}& 87.5 \\
ESIM \cite{DBLP:conf/acl/ChenZLWJI17}& 88.0 \\
DIIN \cite{DBLP:journals/corr/abs-1709-04348}& 88.0 \\
DR-BiLSTM \cite{ghaeini2018dr} & 88.5 \\
CAFE \cite{tay2017compare} & 88.5 \\
\hline
% CSRAN  & 88.2 \\
CSRAN  & \textbf{88.7} \\
\hline
\end{tabular}
\caption{Experimental results on single model SNLI dataset.}
\label{snli}
\end{table}
\vspace{-1em}
\begin{table}[H]
  \centering
\begin{tabular}{lc}
  \hline

Model & Acc \\
\hline
% Majority & 60.3 \\
DecompAtt \cite{DBLP:conf/emnlp/ParikhT0U16} & 72.3 \\
ESIM \cite{DBLP:conf/acl/ChenZLWJI17} & 70.6 \\
% Ngram & 70.6 \\
% DGEM w/o edges & 70.8 \\
DGEM \cite{scitail} & 77.3 \\
CAFE \cite{tay2017compare} & 83.3 \\
\hline
% CSRAN ($l=1$) & 85.8 \\
CSRAN & \textbf{86.7} \\
\hline
\end{tabular}
\caption{Experimental results on SciTail dataset.}
\label{scitail}
\end{table}
\vspace{-1em}

\begin{table}[H]
  \centering
\begin{tabular}{lc}
  \hline

Model & Acc \\
\hline
L.D.C \cite{wang2016sentence} & 87.5 \\
Word DecompAtt \cite{tomar2017neural} & 87.5 \\
BiMPM \cite{1702.03814} & 88.1 \\
Char DecompAtt \cite{tomar2017neural} & 88.4 \\
DIIN \cite{DBLP:journals/corr/abs-1709-04348} & 89.0 \\
\hline
CSRAN & \textbf{89.2} \\
\hline
\end{tabular}
\caption{Experimental results on Quora Duplicate Detection dataset.}
\label{quora}
\end{table}

\vspace{-1em}
\begin{table}[H]
  \centering
\begin{tabular}{lc}
  \hline
Model & F1 \\
\hline
MultiP \cite{xu2014extracting} & 0.536 \\
DeepPairwiseWord \cite{DBLP:conf/naacl/HeL16} & 0.749\\
Subword + LM  \cite{lan2018subword} & 0.760 \\
\hline
CSRAN & \textbf{0.840} \\
\hline
\end{tabular}
\caption{Experimental results on TwitterURL paraphrase dataset.}
\label{twitterurl}
\end{table}
\vspace{-1em}
\begin{table}[H]
  \centering
\begin{tabular}{lc}
  \hline
Model & MAP/MRR \\
\hline
HyperQA \cite{DBLP:journals/corr/TayLH17a} & 0.784/0.865\\
MPCNN \cite{DBLP:conf/cikm/RaoHL16} & 0.801/0.877 \\
BiMPM \cite{1702.03814} & 0.802/0.899\\
CA \cite{Bian:2017:CMD:3132847.3133089} & 0.821/0.899 \\
IWAN \cite{shen2017inter} &	0.822/0.889 \\
MCAN \cite{Tay:2018:MAN:3219819.3220048} & 0.838/0.904  \\
\hline
CSRAN & \textbf{0.854/0.935}\\
\hline
\end{tabular}
\caption{Experimental results on TrecQA dataset.}
\label{trecqa}
\end{table}
\begin{table}[H]
  \centering
\begin{tabular}{lcccc}
  \hline
Model & Acc & R@1 & R@2 & R@5 \\
\hline
CNTN & 0.743 & 0.349 & 0.512 & 0.797 \\
LSTM & 0.725 & 0.361 & 0.494 & 0.801 \\
APLSTM & 0.758 & 0.381 & 0.545 & 0.801 \\
MV-LSTM & 0.767 & 0.410 & 0.565 & 0.800 \\
KEHNN & 0.786 & 0.460 & 0.591 & 0.819 \\
% HCRN & 0.816 & 0.508  & 0.656 & 0.863 \\
MCAN & 0.834 & 0.551 & 0.684 & 0.875 \\
\hline
CSRAN & \textbf{0.839} & \textbf{0.556} & \textbf{0.692} & \textbf{0.880} \\
\hline
\end{tabular}
\caption{Experimental results on the Ubuntu dataset for utterance-response matching. Baseline results are reported from \cite{wu2016knowledge}.}
\label{ubuntu}
\end{table}

\subsubsection{Training Efficency}
With many BiLSTM layers, it is natural to be skeptical about the training efficiency of our model. However, since we use the \textsc{cudnn} implementation of the BiLSTM model, the runtime is actually very manageable. On SNLI, with a batch size of $128$, our model with $3$ stacked recurrent layers and $2$ aggregation BiLSTM layers runs at $\approx$17 minutes per epoch and converges in less than 20 epochs. On SciTail, our model runs at $\approx2$ minutes per epoch with a batch size of $32$. This is benchmarked on a TitanXP GPU. While our model is targetted at performance and not efficiency, this section serves as a reassurance that our model is not computationally prohibitive.
\subsection{Ablation Study}
In order to study the effectiveness of the key components in our proposed architecture, we conduct an extensive ablation study. Table \ref{tab:ablation}
 reports the results on several ablation baselines. There are three key ablation baselines as follows: (1) we removed MAR from the stacked recurrent
 network, (2) we removed CSRA from the network and finally (3) we removed both MAR and CSRA from the network. All ablation baselines reported are stacked with $3$ layers.
 % Table generated by Excel2LaTeX from sheet 'Sheet2'
\begin{table}[htbp]

  \centering

    \begin{tabular}{lccc}
      \hline
    Ablation & SNLI &SciTail & TrecQA \\
    \hline
    Original & 88.6 & 88.0 & 0.86/0.90 \\
    w/o MAR & 88.4 & 82.5 & 0.79/0.85 \\
    w/o CSRA & 88.1 &86.2 & 0.84/0.89 \\
    w/o MAR/CSRA & 88.0 & 83.0    & 0.79/0.84 \\
    \hline
    \end{tabular}%
      \caption{Ablation study (development score) of our key model components on three datasets. }
  \label{tab:ablation}%
\end{table}%
Firstly, we observe that both MAR and CSRA are critical components in our model, i.e., removing any of them would result in a drop in performance. Secondly, we observe that the relative utility of CSRA and MAR depends on the dataset. Removing MAR sigificantly reduces performance on SciTail and TrecQA. On the other hand, removing CSRA degrades the performance more than MAR on SNLI. Finally, it is good to note that, while performance degradation on SNLI development set may not seem significant, the \textit{w/o MAR and CSRA} ablation performance baseline achieved only $87.7\%$ accuracy on the test set, compared to $88.7\%$ of the original model. This is equivalent to dropping from state-of-the-art to the 5th ranked model. Overall, we are able to conclude that the CSRA and MAR make meaningful improvements to our model architecture.

\subsection{Effect of Stack Depth}

In this section, our goals are twofold - (1) studying the effect of stack depth on model performance and (2) determining if the proposed CSRAN model indeed helps with enabling deeper stack depths.
In order to do so, we compute the development set performance of two models. The first is the full CSRAN architecture and the second is a baseline stacked model architecture. Note that the bidirectional alignment layer and remainder of the model architecture (highway layers, etc.) remain completely identical to CSRAN to make this study as fair as possible.
\begin{figure}[H]
  \centering
    \includegraphics[width=0.8\linewidth]{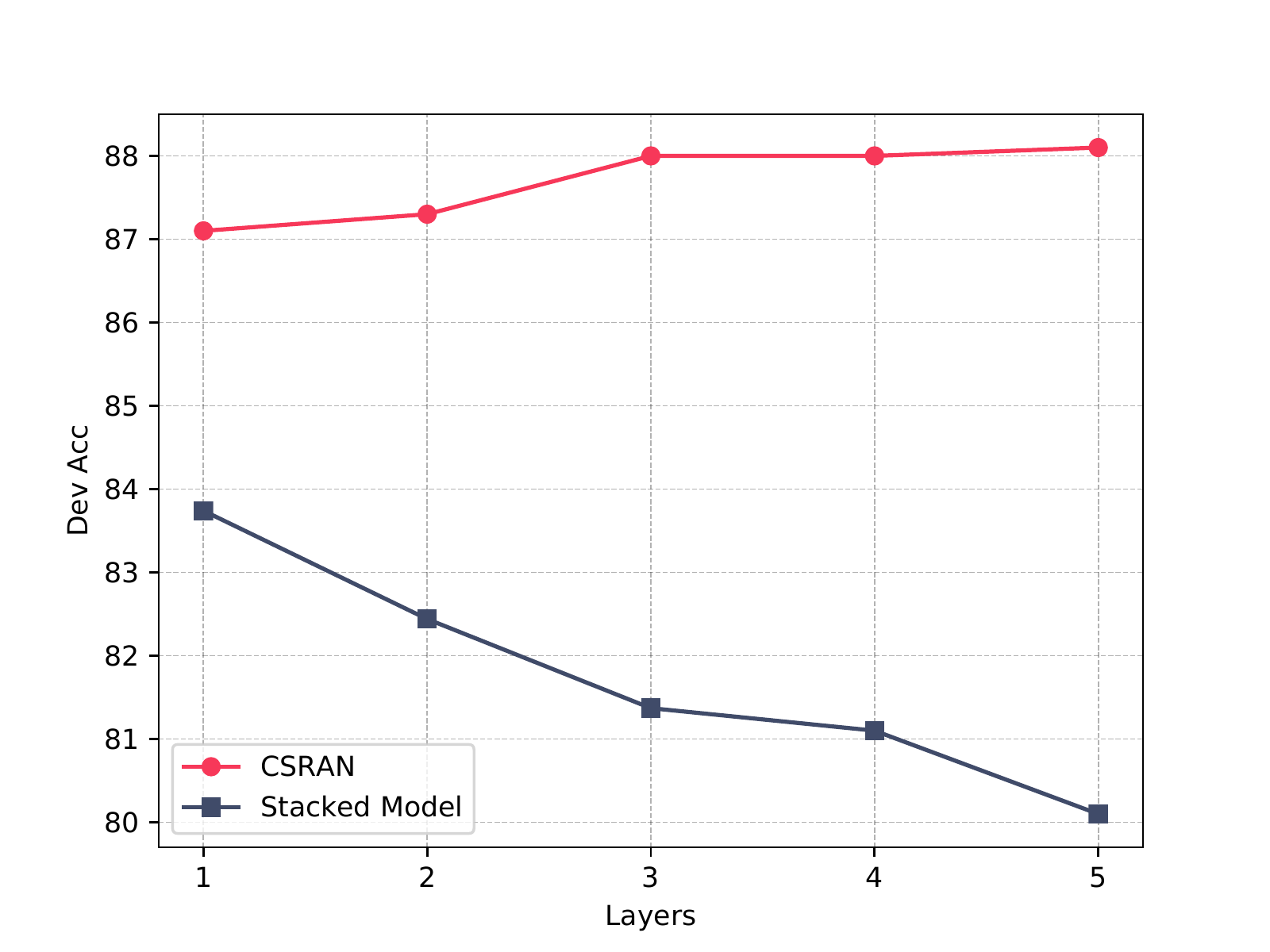}
    \caption{Relative effect of stack depth on CSRAN and the baseline Stacked Model on SciTail dataset.}
    \label{fig:scitail_layer}
\end{figure}

Figure \ref{fig:scitail_layer} illustrates the model performance with varying stack depth. As expected, the performance of the stacked model declines when increasing the stack depth. On the other hand, the performance of CSRAN improves by adding additional layers. The largest gain is when jumping from 2 layers to 3 layers. The subsequent performance improvement from 3-5 layers is marginal. From this study, the takeaway is that standard stacked architectures are insufficient. As such, our proposed CSRA mechanism can aid in enabling deeper models which can result in stronger model performance\footnote{The best result on Scitail was obtained with 5 layers. Moreover, the difference in test performance between stacked and single-layered model was considerably high ($+2.5\%$) even though dev performance increased by $+1\%$. }.

\begin{figure}[H]
  \centering
    \includegraphics[width=0.8\linewidth, height=4.2cm]{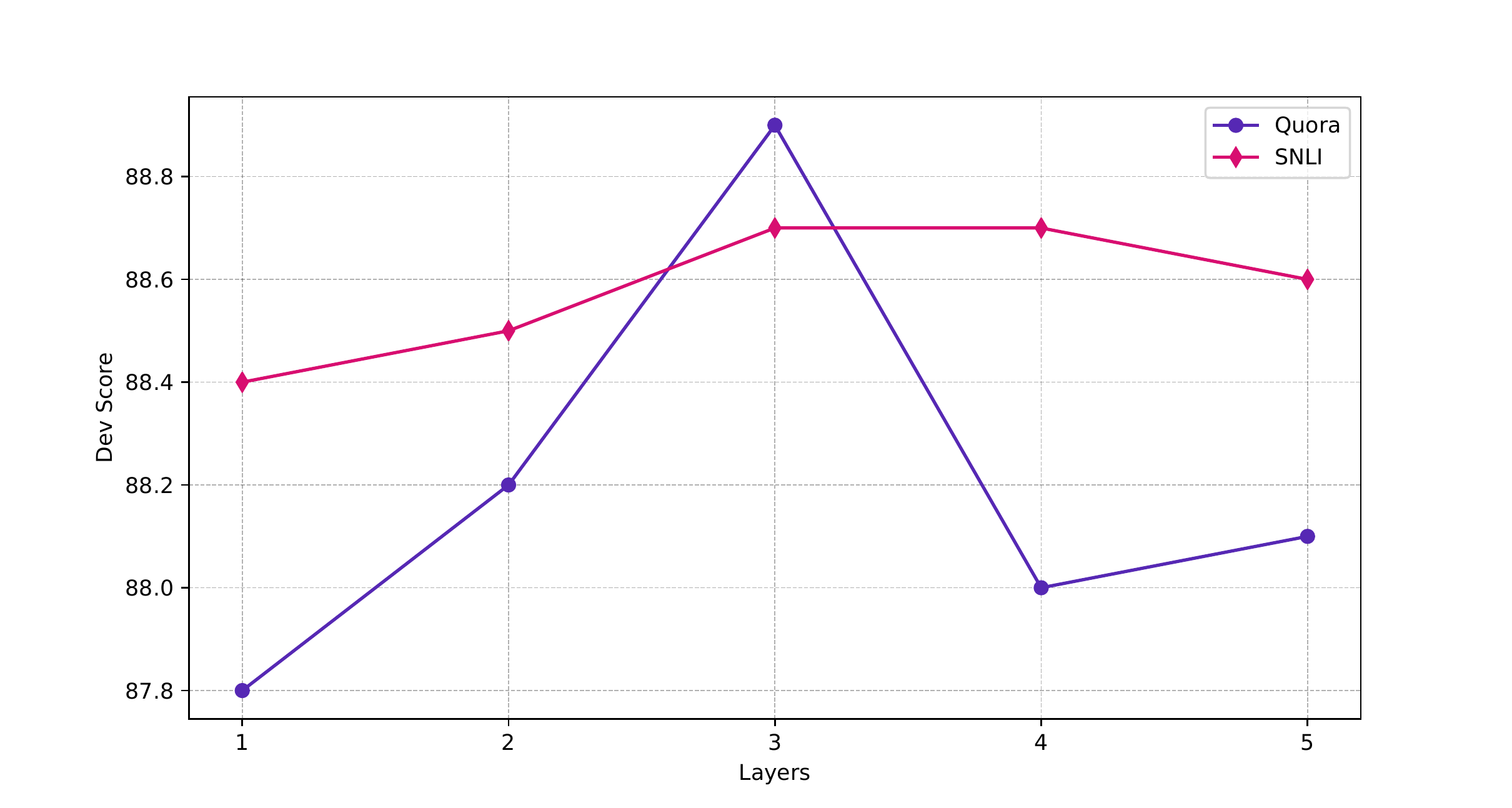}
    \caption{Effect of stack depth on CSRAN performance on Quora and SNLI datasets.}
    \label{fig:snli_layer}
\end{figure}

Next, we study the general effect of stack depth (number of layers) on model performance. Figure \ref{fig:snli_layer} reports the model performance (dev accuracy) of our CSRAN architecture on Quora and SNLI datasets. We observe that a stacked architecture with 3 layers is significantly better than a single-layered architecture. The optimal development score is 3-4 layers for SNLI and 3 layers for Quora. However, we observe the performance of Quora declines after 3 layers (notably it is still higher than an unstacked model). However, the performance on SNLI remains relatively stable.

\section{Related Work}
Learning to matching text sequences is a core and fundamental research problem in NLP and Information Retrieval. A wide range of NLP applications fall under this paradigm such as natural language inference \cite{DBLP:conf/emnlp/BowmanAPM15,scitail}, paraphrase identification \cite{lan2018subword}, question answering \cite{DBLP:conf/sigir/SeverynM15}, document search \cite{shen2014latent,hui2017pacrr}, social media search \cite{rao2018multi} and entity linking \cite{phan2017neupl}. As such, universal text matching algorithms are generally very attractive, in lieu of the prospects of potentially benefitting an entire suite of NLP applications.

Neural networks have been the prominent choice for text matching. Earlier works are mainly concerned with learning a matching function between RNN/CNN encoded representations \cite{DBLP:conf/sigir/SeverynM15,DBLP:journals/corr/YuHBP14,DBLP:conf/ijcai/QiuH15,DBLP:conf/sigir/TayPLH17,1711.07656}. Models such as Recursive Neural Networks have also been explored \cite{wan2016match}. Subsequently, attention-based models were adopted \cite{rocktaschel2015reasoning,DBLP:conf/acl/WangL016,DBLP:conf/emnlp/ParikhT0U16}, demonstrating superior performance relative to their non-attentive counterparts.

Today, the dominant state-of-the-art approaches for text matching are mostly based on neural models configured with bidirectional attention layers \cite{shen2017inter,tay2017compare}. Bidirectional attention comes in various flavours which can be known as soft alignment \cite{shen2017inter,DBLP:conf/acl/ChenZLWJI17}, decomposable attention \cite{DBLP:conf/emnlp/ParikhT0U16}, attentive pooling \cite{DBLP:journals/corr/SantosTXZ16} and even complex-valued attention \cite{tay2018hermitian}. The key idea is to jointly soft align text sequences such that they can be compared at the index level. To this end, various comparison functions have been utilized, ranging from feed-forward neural networks \cite{DBLP:conf/emnlp/ParikhT0U16} to factorization machines \cite{tay2017compare}. Notably, these attention (and bi-attention) mechanisms are also widely adopted (or originated) from many related sub-fields of NLP such as machine translation \cite{bahdanau2014neural} and reading comprehension \cite{DBLP:journals/corr/XiongZS16,seo2016bidirectional,wang2016machine}.

Many text matching neural models are heavily grounded in the compare-aggregate architecture \cite{DBLP:journals/corr/WangJ16b}. In these models, matching and comparisons occur between text sequences, aggregating features for making the final prediction. Recent state-of-the-art models such as BiMPM \cite{1702.03814} and DIIN \cite{DBLP:journals/corr/abs-1709-04348} are representative of such architectural paradigm, utilizing an attention-based matching scheme and then a CNN or LSTM-based feature aggregator. Earlier works \cite{DBLP:conf/aaai/WanLGXPC16,DBLP:conf/emnlp/HeGL15,DBLP:conf/naacl/HeL16} exploit a similar paradigm, albeit without the usage of attention.

Across many NLP and machine learning applications, utilizing stacked architectures is a common way to enhance representation capability of the encoder \cite{sutskever2014sequence,graves2013speech,zhang2016highway,DBLP:conf/repeval/NieB17}, leading to performance improvement. Deep networks suffer from inherent difficulty in feature propagation and/or vanishing/exploding gradients. As a result, residual strategies have often been employed \cite{he2016deep,DBLP:journals/corr/SrivastavaGS15,DBLP:conf/cvpr/HuangLMW17}. However, to the best of our knowledge, this work presents a new way of residual connections, leveraging on the fact that pairwise formulation of the text matching task.

\section{Conclusion}
We proposed a deep stacked recurrent architecture for general-purpose text sequence matching. We proposed a new co-stack residual affinity mechanism for matching sequence pairs, leveraging multi-hierarchical information for learning bidirectional alignments. Our proposed CSRAN model achieves state-of-the-art performance across six well-studied benchmark datasets and four different problem domains.

\bibliography{emnlp2018}
\bibliographystyle{acl_natbib_nourl}

\end{document}